\documentclass[sigconf,natbib=true,anonymous=false]{acmart}
\usepackage{booktabs}
\usepackage{multirow}
\usepackage{tabularx}
\usepackage{array}
\usepackage{float}
\usepackage{subcaption}
\usepackage[table]{xcolor}
\definecolor{lightpurple}{RGB}{235,230,250}
\newcolumntype{Y}{>{\centering\arraybackslash}X}

\AtBeginDocument{%
  }

\setcopyright{acmlicensed}
\copyrightyear{2026}
\acmYear{2026}
\acmDOI{XXXXXXX.XXXXXXX}
\acmConference[ SIGIR '26]{Make sure to enter the correct
  conference title from your rights confirmation email}{June 03--05,
  2026}{Melbourne, Australia}
\acmISBN{XXX-X-XXXX-XXXX-X/2026/06}
\begin{document}

\title{Diff-SBSR: Learning Multimodal Feature-Enhanced Diffusion Models for Zero-Shot Sketch-Based 3D Shape Retrieval}

\author{Hang Cheng}
\authornote{Both authors contributed equally to this research.}
\affiliation{%
  \institution{Tsinghua Shenzhen International Graduate School, Tsinghua University}
  \city{Shenzhen}
  \state{Guangdong}
  \country{China}
}
\email{chenghan24@mails.tsinghua.edu.cn}

\author{Fanhe Dong}
\authornotemark[1]
\affiliation{%
  \institution{Tsinghua Shenzhen International Graduate School, Tsinghua University}
  \city{Shenzhen}
  \state{Guangdong}
  \country{China}}
\email{dfh25@mails.tsinghua.edu.cn}

\author{Long Zeng}
\authornote{Corresponding author.}
\affiliation{%
  \institution{Tsinghua Shenzhen International Graduate School, Tsinghua University}
  \city{Shenzhen}
  \state{Guangdong}
  \country{China}}
\email{zenglong@sz.tsinghua.edu.cn}


\begin{abstract}
This paper presents the first exploration of text-to-image diffusion models for zero-shot sketch-based 3D shape retrieval (ZS-SBSR). Existing sketch-based 3D shape retrieval methods struggle in zero-shot settings due to the absence of category supervision and the extreme sparsity of sketch inputs. Our key insight is that large-scale pretrained diffusion models inherently exhibit open-vocabulary capability and strong shape bias, making them well suited for zero-shot visual retrieval. We leverage a frozen Stable Diffusion backbone to extract and aggregate discriminative representations from intermediate U-Net layers for both sketches and rendered 3D views. Diffusion models struggle with sketches due to their extreme abstraction and sparsity, compounded by a significant domain gap from natural images. To address this limitation without costly retraining, we introduce a multimodal feature-enhanced strategy that conditions the frozen diffusion backbone with complementary visual and textual cues from CLIP, explicitly enhancing the ability of semantic context capture and concentrating on sketch contours. Specifically, we inject global and local visual features derived from a pretrained CLIP visual encoder, and incorporate enriched textual guidance by combining learnable soft prompts with hard textual descriptions generated by BLIP. Furthermore, we employ the Circle-T loss to dynamically strengthen positive-pair attraction once negative samples are sufficiently separated, thereby adapting to sketch noise and enabling more effective sketch-3D alignment. Extensive experiments on two public benchmarks demonstrate that our method consistently outperforms state-of-the-art approaches in ZS-SBSR.
\end{abstract}



\maketitle

\section{Introduction}
Notably, the rapid growth of 3D shape data in recent years has significantly advanced research on 3D shape retrieval, making it a core topic in multimedia retrieval \cite{Xie2016TPAMI}. With the widespread adoption of touch-screen devices, freehand sketches have emerged as a convenient and intuitive form of human–computer interaction. Compared to textual queries or direct manipulation of 3D models, sketches offer a more natural and accessible way for users to express visual intent. However, unlike natural images, sketches are characterized by extreme abstraction and sparsity \cite{Bandyopadhyay2024CVPRExplain, Bandyopadhyay2024CVPRSketchINR, zeng2019sketch}, and their cross-modal nature makes it non-trivial to learn reliable representations. Consequently, sketch-based 3D shape retrieval (SBSR) presents a challenging yet practically relevant retrieval problem \cite{bai2025scdl, su2025dkd}.

To the best of our knowledge, the domain-disentanglement GAN (DD-GAN) proposed in \cite{Xu2022AAAI} is the first work to explicitly address zero-shot learning in the SBSR setting. DD-GAN decomposes features into domain-invariant and domain-specific components, and enables knowledge transfer from seen to unseen classes by aligning domain-invariant features with their corresponding word embeddings. Building on this line of research, subsequent methods have further advanced zero-shot SBSR. D2O \cite{Wang2023AAAI}introduces prototype contrastive learning, using class prototypes as intermediates to align channel-wise features across modalities. Codi \cite{Meng2024TCSVT} further improves zero-shot transfer by directly enforcing semantic alignment between domain-invariant features and word embeddings, leading to improved retrieval performance. Despite these advances, most existing zero-shot SBSR methods rely on explicit semantic alignment or prototype-based supervision, and fail to adequately handle the extreme sparsity and subjective variability of sketches, limiting generalization to unseen categories.

More recently, with the rapid development of large-scale pretrained vision-language models \cite{Li2022ICML} and text-to-image generative models \cite{dhariwal2021diffusion, rombach2022high}, a natural direction has emerged: leveraging their rich semantic priors and open-vocabulary properties to support zero-shot generalization to unseen categories \cite{baranchuk2021label, hu2020sketchsegmenter}. In particular, text-to-image diffusion models have been shown to effectively bridge the modality gap between sketches and photos, benefiting from strong cross-modal alignment and an inherent shape bias, as also observed in recent work \cite{koley2025sketchfusion, koley2024text}. Although Stable Diffusion (SD) has demonstrated impressive feature extraction capabilities across a wide range of vision tasks, its applicability to more challenging sketch-centric cross-modal scenarios remains underexplored. Unlike pixel-rich photographs, sketches are binarized, extremely sparse, and convey limited semantic cues; consequently, diffusion features extracted from sketches tend to be less structurally consistent and semantically explicit than those derived from natural images. This limitation arises from two main factors: (i) the abstract nature and absence of texture or color make it difficult for SD to capture fine-grained visual cues from sketches; and (ii) SD is pretrained exclusively on natural images, which hampers semantic inference when applied to an extremely sparse modality. Moreover, diffusion models encapsulate a diverse set of features across different layers and timesteps, which must be carefully selected and adapted to suit specific downstream tasks.

\begin{figure}[t]
  \centering
  \includegraphics[
    width=0.8\linewidth,
    trim=0mm 0mm 0mm 0mm,
    clip
]{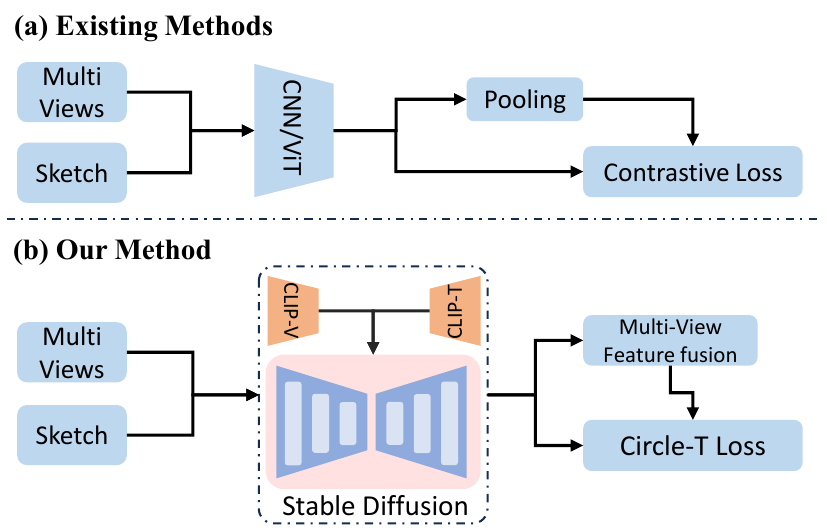}
  \caption{Existing method and our method.}
  \Description{Compared with existing method and our method.}
\end{figure}

To overcome these challenges, we introduce Diff-SBSR, a feature-enhanced diffusion framework for zero-shot sketch-based 3D shape retrieval, which improves robustness to sketch sparsity and unseen classes. Specifically, we render each 3D shape into multi-view 2D images offline, and feed these views together with the input sketch into a shared-weight SD feature extractior. We register forward hooks on multiple down-sampling and up-sampling blocks along the forward denoising trajectory to obtain multi-scale intermediate features. For feature aggregation, we map multi-scale features from different U-Net layers into a unified dimensionality and obtain compact single-image representations via global pooling and learnable weighted fusion. We further identify the layers and timesteps that provide the most informative features for SBSR, and optimize/stabilize the interaction with Stable Diffusion to enable effective conditional feature extraction.

To further strengthen semantic discriminability and cross-modal consistency, we perform multi-modal injection into SD feature extraction from both visual features and textual conditioning. First, we inject local patch-level representations from a pre-trained CLIP visual encoder into SD U-Net intermediate features via channel alignment and residual mapping, enabling the denoising process to explicitly perceive high-level semantic structure and local shape cues. Second, leveraging an IP-Adapter mechanism, we project image-level global semantic vectors into cross-attention tokens and inject them into U-Net cross-attention layers, enhancing the modeling of overall geometric layout. Third, for textual conditioning, we employ BLIP-generated captions as hard prompts and introduce learnable continuous soft prompt embeddings to input CLIP text encoder to mitigate the lack of high-quality textual annotations in sketch and 3D-view data, thus steering diffusion features toward more discriminative directions without explicit category supervision.

For 3D shapes, we first remove redundant views by ranking based on CLIP text embedding distances, then feed the selected views into the SD feature extractor. The multi-view features output by the SD extractor are further fused through cross-view adaptive aggregation to generate the final 3D embedding.We adopt the Circle-T loss to dynamically enhance positive-pair attraction once negatives are sufficiently separated, effectively handling sketch noise and enabling robust sketch-3D alignment. This cosine-similarity-based metric learning establishes a stable cross-modal decision boundary between sketches and 3D views, significantly improving retrieval performance in zero-shot settings.

Our main contributions are as follows.
\begin{itemize}
\item {We present the first exploration of Stable Diffusion for zero-shot sketch-based 3D retrieval. We form discriminative representations by extracting and aggregating intermediate U-Net features, leveraging the open-vocabulary capability and shape bias of SD to handle zero-shot retrieval scenarios.
}
\item {To mitigate sketch sparsity and domain mismatch, we inject global and local visual features and enriched textual prompts (soft + BLIP-generated hard) into the frozen backbone, explicitly enhancing the ability of semantic context capture and concentrating on sketch contours.
}
\item {We introduce a circle-T loss that dynamically strengthens positive-pair attraction after negative separation, reducing the impact of sketch noise and improving sketch–3D alignment in zero-shot retrieval.
}
\item {Extensive experiments on public benchmarks demonstrate that our approach consistently outperforms state-of-the-art methods, validating its effectiveness.
}
\end{itemize}

\section{Related Work}
\noindent\textbf{Sketch-based 3D Shape Retrieval.} Sketch-based 3D shape retrieval (SBSR) is intrinsically challenging due to the large modality gap between abstract 2D sketches and complex 3D geometry. Early approaches relied on hand-crafted sketch features and shallow models \cite{Yoon2010ACMMM, Saavedra2012EG3DOR, Li2013SHREC, Li2014CVIU, Li2014SHREC}, which were limited in capturing cross-modal semantics. Deep learning significantly advanced SBSR by enabling nonlinear representation learning \cite{Wang2015CVPR, Dai2018TIP, Chen2018ECCV, Qin2022CAG}, leading to a dominant paradigm that projects sketches and 3D shapes into a shared embedding space via metric learning \cite{Lei2019PR, Dai2017AAAI, Dai2018TIP}. For 3D representation, early model-based methods employed voxels, point clouds, or dedicated 3D encoders \cite{maturana2015voxnet, klokov2017escape, Qi2018BMVC, wang2017cnn, esteves2018learning, qi2017pointnet, qi2017pointnet++}, but their high computational cost limits scalability. View-based methods mitigate this issue by representing 3D shapes through multiple 2D projections \cite{Wang2015CVPR, yu2016sketch, darom2012scale, Dai2017AAAI}, yet aligning sparse sketches with multi-view representations remains inherently ambiguous. To alleviate this, recent works introduce attention mechanisms and cross-modal interaction strategies \cite{bai2023pagml, xu2020tmm, zhou2020tvcg, Zhao2022JVCIR}, including cross-modal view attention \cite{Qi2021TIP} and teacher–student alignment frameworks \cite{Dai2020ICME,Liang2024CVIU}. Most view-based approaches still rely on global aggregation over shared CNN features \cite{Dai2020ICME,Cai2023ICME,Liang2024CVIU, su2015multi}, which can obscure inter-view relationships, motivating recent explorations of Transformer-based aggregation and viewpoint selection \cite{zhu2024sketch, zhang2025meha, yuan2023retrieval}. In parallel, uncertainty modeling and stroke-level sketch representation have been investigated to handle sketch sparsity and ambiguity \cite{Liang2021TIP, Cai2023ICME, Bai2023KBS,su2025dkd,su2025skd}.

\noindent\textbf{Diffusion-based backbone for Downstream Tasks.} Diffusion models, formulated as a progressive denoising process, have become a dominant paradigm for high-fidelity visual synthesis, achieving strong performance across 2D image generation~\cite{dhariwal2021diffusion, rombach2022high}, 3D content creation~\cite{chou2023diffusion, bandyopadhyay2024doodle}, and video generation~\cite{chen2024videocrafter2, ho2020ddpm}. Their powerful text-conditioned modeling has further enabled a range of image editing and personalization methods, including Prompt-to-Prompt~\cite{hertz2022prompt}, Imagic~\cite{kawar2023imagic}, and DreamBooth~\cite{ruiz2023dreambooth}. Beyond generative applications, recent studies have demonstrated that diffusion models learn rich intermediate representations with strong semantic and structural priors, making them effective feature extractors for discriminative and dense prediction tasks, such as retrieval~\cite{sain2023sdpl}, classification and segmentation~\cite{baranchuk2021label, hu2020sketchsegmenter}, depth estimation~\cite{baranchuk2021label}, object detection~\cite{chowdhury2023detection}, image translation and controllable generation~\cite{ham2022cogs, bandyopadhyay2024doodle}, as well as medical imaging~\cite{dewilde2024medical}. To further enhance task performance, several works introduce task-specific diffusion variants through test-time adaptation~\cite{wang2023tta}, masked reconstruction~\cite{chen2023masked}, or compact latent modeling~\cite{hudson2024soda}. However, these approaches often rely on carefully designed mechanisms and exhibit limited robustness under domain shifts. To address the difficulty of extracting discriminative representations from highly sparse sketches, we adopt a multimodal feature enhancement strategy to augment pretrained diffusion-based feature extraction, thereby providing more robust representations for ZS-SBSR.

\noindent\textbf{Zero-shot Sketch-based Retrieval.} Zero-shot sketch-based retrieval aims to generalize retrieval models to categories unseen during training, placing strong demands on cross-modal generalization. Existing approaches typically rely on auxiliary semantic supervision, attribute-level modeling, or pretrained vision–language representations to mitigate this challenge \cite{sain2023clip, lin2023zero, li2024dr, singha2024elevating}. In contrast, zero-shot sketch-based image retrieval has been more extensively studied, with a range of solutions including conditional generation \cite{yelamarthi2018zero}, cross-modal reconstruction \cite{deng2020progressive}, relationship-aware knowledge distillation \cite{tian2021relationship}, and test-time adaptation strategies \cite{sain2022sketch3t}. Compared to image-based retrieval, zero-shot SBSR remains relatively underexplored. DDGAN \cite{Xu2022AAAI} represents an early attempt to extend SBSR to the zero-shot setting through adversarial learning with disentangled representations. Subsequent works introduced prototype-based learning with stroke-level sketch modeling \cite{Wang2023AAAI}, leveraged images as an intermediate modality to acquire adaptive 3D viewpoint knowledge \cite{Chowdhury2023ICCV}, and further enhanced semantic transfer by incorporating semantic prototypes \cite{Meng2024TCSVT}.

\section{Method}

\subsection{Problem Statement}
Given a query sketch $\mathbf{s}$ belonging to a specific class $y$, Zero-Shot Sketch-Based 3D Shape Retrieval (ZS-SBSR) aims to learn a transferable model that retrieves a semantically relevant \textit{3D shape} $\mathcal{M}$ from a gallery $\mathcal{G} = \{\mathcal{M}_i\}_{i=1}^{N_G}$.
The gallery contains $N_G$ 3D shapes spanning multiple categories, where a retrieval is considered successful if the retrieved shape shares the same label as the query (i.e., $y_{\mathcal{M}} = y_{\mathbf{s}}$).
Unlike conventional SBSR which evaluates on categories \textit{seen} during training ($\mathcal{C}^S$), the ZS paradigm focuses on evaluating on \textit{mutually exclusive} \textit{unseen} categories $\mathcal{C}^U = \{c_1^U, \dots, c_M^U\}$, such that $\mathcal{C}^S \cap \mathcal{C}^U = \emptyset$.

This work presents a feature-enhanced diffusion models for zero-shot sketch-based 3D shape retrieval.

\begin{figure*}[t]
  \centering
  \includegraphics[
    width=0.8\textwidth,
    trim=0mm 2mm 0mm 5mm,
    clip
  ]{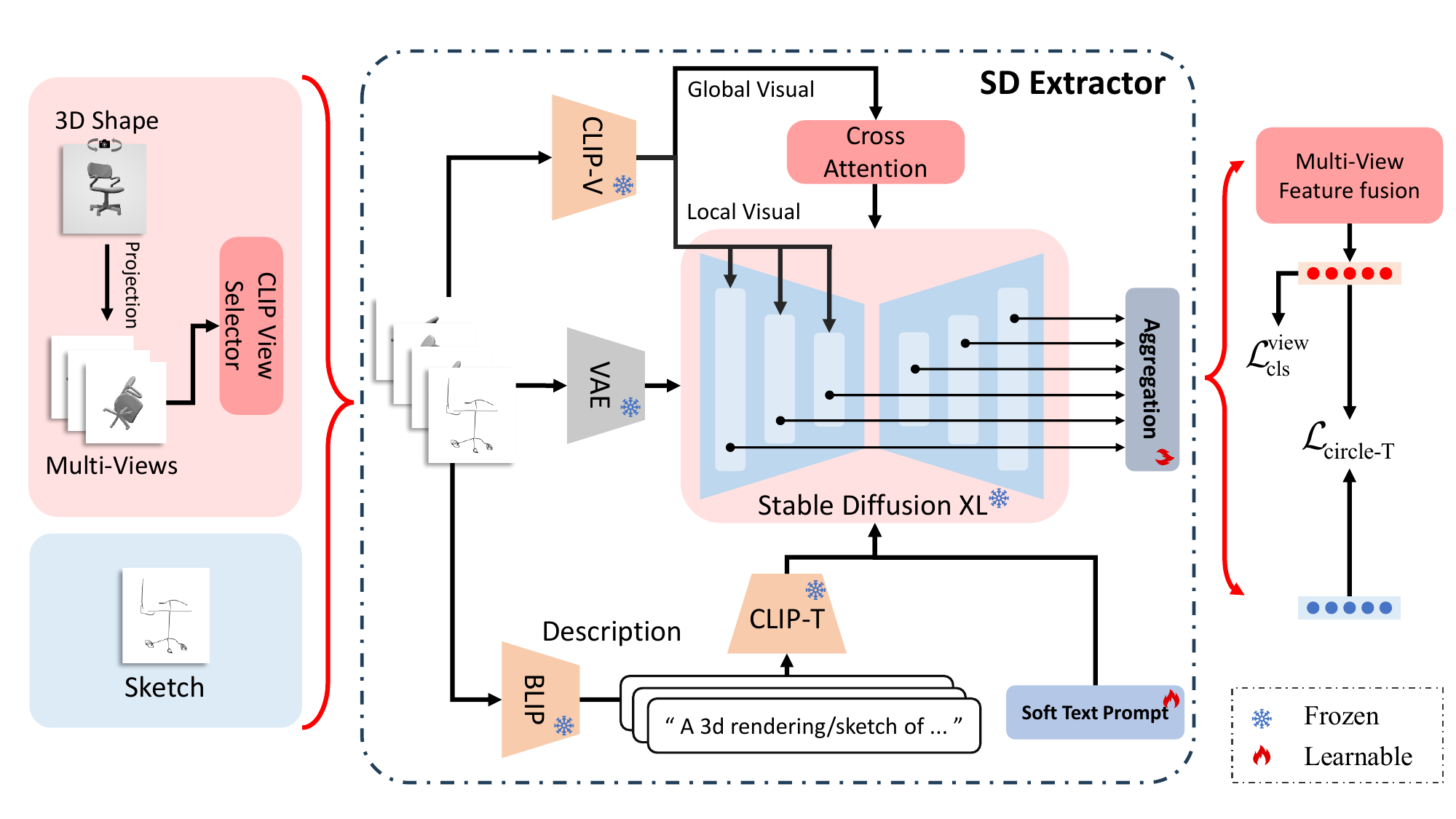}
  \caption{Our method extracts multi-scale features from a frozen Stable Diffusion model for sketches and rendered 3D views, and enhances them with visual and textual conditioning derived from pretrained vision–language models. The conditioned features are aggregated across views to obtain final embeddings for zero-shot sketch-based 3D shape retrieval.}
  \Description{Our method extracts multi-scale features from a frozen Stable Diffusion model for sketches and rendered 3D views, and enhances them with visual and textual conditioning derived from pretrained vision–language models. The conditioned features are aggregated across views to obtain final embeddings for zero-shot sketch-based 3D shape retrieval.}
\end{figure*}

\subsection{Diffusion model Preliminary}
 Diffusion Model generates images by progressively removing noise from an initial two-dimensional isotropic Gaussian distribution through an iterative refinement process. This process consists of two complementary procedures: a \emph{forward} diffusion process and a \emph{reverse} denoising process~\cite{ho2020ddpm, rombach2022high}. 

In the forward process, random Gaussian noise is gradually added to a clean image $i_0 \in \mathbb{R}^{h \times w \times 3}$ over $t$ timesteps, yielding a noisy image $i_t$:
\begin{equation}
    i_t = \sqrt{\bar{\alpha}_t} \, i_0 + \sqrt{1 - \bar{\alpha}_t} \, \epsilon,
\end{equation}
where $\epsilon \sim \mathcal{N}(0, I)$ denotes Gaussian noise, $\{\alpha_t\}_{t=1}^T$ is a predefined noise schedule with $\bar{\alpha}_t = \prod_{k=1}^t \alpha_k$, and $t$ is uniformly sampled from $\{1, \dots, T\}$. 

In the reverse process, a denoising U-Net $U_\theta$ is trained with an $\ell_2$ objective to predict the added noise from the noisy input, i.e., $\epsilon \approx U_\theta(i_t, t)$. After training, $U_\theta$ can reconstruct clean images from noisy inputs. During inference, the process starts from a random noise sample $i_T \sim \mathcal{N}(0, I)$, and $U_\theta$ is applied iteratively for $T$ steps to gradually reduce noise and generate progressively cleaner intermediate images $i_{t-1}$, ultimately producing a sample $i_0$ from the target distribution. By incorporating auxiliary conditioning signals such as text captions $c$, the model can be extended to conditional generation, where $U_\theta(i_t, t, c)$ performs denoising under the guidance of $c$, typically via cross-attention.

Pixel-space diffusion models are computationally expensive as they operate directly on high-resolution images ($i_0 \in \mathbb{R}^{h \times w \times 3}$). In contrast, Stable Diffusion (SD) model performs diffusion in a lower-dimensional latent space, resulting in significantly improved efficiency and stability. Specifically, SD is trained in two stages. In the first stage, a variational autoencoder (VAE) with encoder $E(\cdot)$ and decoder $D(\cdot)$ is learned, mapping an input image to a latent representation $z_0 = E(i_0) \in \mathbb{R}^{\frac{h}{8} \times \frac{w}{8} \times d}$. In the second stage, a U-Net $U_\theta$ is trained to perform denoising directly in this latent space, conditioned on text embeddings generated by a text encoder $T(\cdot)$. The training objective is:
\begin{equation}
    \mathcal{L}_{\ell_2} = \mathbb{E}_{z_t, t, c, \epsilon} \left[ \left\| \epsilon - U_\theta(z_t, t, T(c)) \right\|_2^2 \right].
\end{equation}
During inference, a noisy latent $z_t$ is initialized as $z_t \sim \mathcal{N}(0, I)$, and $U_\theta$ is applied iteratively over $T$ timesteps conditioned on the composite prompt embeddings $c$ to produce a denoised latent $\hat{z}_0 \in \mathbb{R}^{\frac{h}{8} \times \frac{w}{8} \times d}$. The final high-resolution image is then reconstructed as $\hat{i} = D(\hat{z}_0) \in \mathbb{R}^{h \times w \times 3}$.

\subsection{Overview}

In response to the significant modal differences between sketches and 3D objects, we propose an enhanced SD feature extraction method that integrates visual cues and textual prompts.

Firstly, to construct an effective representation of 3D objects, we adopt a strategy of multi-view projection combined with CLIP-guided view selection. Specifically, we render 3D objects from several fixed viewpoints to obtain a set of candidate projection images $\{v_i\}$. To reduce data redundancy while retaining key semantics, we use the pre-trained CLIP image encoder to calculate the cosine similarity between the visual embedding $\mathbf{f}_i = \mathrm{CLIP}_{\mathrm{img}}(v_i)$ of each view and the text embedding $\mathbf{e}_{txt}$ derived from the description \textbf{\textit{``a photo of \{class\}.''}}. For each object, we select the top-$k$ views $\mathcal{V}^{*}$ with the highest similarity scores:
\begin{equation}
\mathcal{V}^{*} = \operatorname{TopK}_{k}\big(\cos(\mathbf{f}_i, \mathbf{e}_{txt})\big), \quad k=3.
\end{equation}

Secondly, in the core feature extraction stage, relying solely on the pre-trained weights of SD proves insufficient to precisely align sketches and 3D renderings due to their modal gap.  Therefore, we utilize SD as a powerful feature extraction backbone, augmented by a hybrid injection mechanism. We specifically employ Stable Diffusion XL (SDXL) in our work, as it features a significantly larger U-Net and a dual-text encoder architecture. SDXL utilizes two frozen text encoders: CLIP ViT-L ($T_L$) and OpenCLIP ViT-bigG ($T_G$). The token features from $T_L$ and $T_G$ are concatenated to form a rich context for cross-attention, while the pooled output features of $T_G$ are injected into the time-step embedding to control the global style. For an input image $x$ (which can be a sketch $s$ or a selected view $v \in \mathcal{V}^{*}$), we feed it into the SD encoder. To enhance the representation, we inject CLIP visual features—specifically the global CLS token $\mathbf{c}_{cls}$ and local patch tokens $\mathbf{c}_{patch}$—along with soft/hard textual prompts ($\mathbf{p}_{soft}, \mathbf{p}_{hard}$). This process produces the final feature representation $\mathbf{F}$, formally expressed as:
\begin{equation}
\mathbf{F} = \mathrm{SD}(x, \mathbf{c}_{cls}, \mathbf{c}_{patch}, \mathbf{p}_{soft}, \mathbf{p}_{hard}).
\end{equation}
This mechanism ensures that the extracted features are both high-quality and discriminative (Sec.~\ref{sec:visual_enhancement} and Sec.~\ref{sec:textual_injection}).

Furthermore, in our experiments, we observed that the commonly used quadruplet loss often struggles to dynamically balance the relative distances of positive and negative pairs. To address this, we introduce dynamic adjustment parameters based on the Circle Loss (Sec.~\ref{sec:Circle-T}). This improved objective function flexibly adjusts the optimization trajectory, ensuring inter-class separability while maintaining intra-class compactness.

Finally, to obtain a unified metric representation, we aggregate the features. For a 3D object, the global representation $\mathbf{F}_{3D}$ is obtained by performing max pooling over the features of the selected views. For a sketch, its feature $\mathbf{F}_s$ is used directly:
\begin{equation}
\mathbf{F}_{3D} = \operatorname{MaxPool}_{v \in \mathcal{V}^{*}} \big( \bar{\mathbf{f}}_v \big), \quad \mathbf{F}_{sketch} = \bar{\mathbf{f}}_s.
\end{equation}

\subsection{Feature Extraction via Stable Diffusion}
The internal activations of a pretrained SDXL UNet $U_{\theta}$ contain informative representations.
Given an input image $\mathbf{i}$, it is first encoded by a pretrained VAE encoder $E(\cdot)$ into an initial latent representation $\mathbf{z}_0 = E(\mathbf{i}) \in \mathbb{R}^{\frac{H}{8} \times \frac{W}{8} \times d}$.
At a given diffusion timestep $t$, SDXL performs forward diffusion on $\mathbf{z}_0$ to obtain the noisy latent representation $\mathbf{z}_t$.
Then, $\mathbf{z}_t$, the timestep $t$, and the conditioning embeddings (detailed in Sec.~\ref{sec:visual_enhancement} and Sec.~\ref{sec:textual_injection}) are fed into the pretrained denoising UNet $U_{\theta}(\cdot)$.

We extract internal feature maps from both the downsampling and upsampling layers of the UNet via hook mechanisms.
For SDXL, given an input image $\mathbf{i} \in \mathbb{R}^{H \times W \times 3}$, the downsampling layers $\{U_n^{d}\}_{n=1}^{3}$ produce feature maps
$\mathbf{f}_1^{d} \in \mathbb{R}^{\frac{H}{8} \times \frac{W}{8} \times 320}$,
$\mathbf{f}_2^{d} \in \mathbb{R}^{\frac{H}{16} \times \frac{W}{16} \times 640}$,
and
$\mathbf{f}_3^{d} \in \mathbb{R}^{\frac{H}{32} \times \frac{W}{32} \times 1280}$.
Correspondingly, the upsampling layers $\{U_n^{u}\}_{n=1}^{3}$ yield feature maps
$\mathbf{f}_1^{u} \in \mathbb{R}^{\frac{H}{32} \times \frac{W}{32} \times 1280}$,
$\mathbf{f}_2^{u} \in \mathbb{R}^{\frac{H}{16} \times \frac{W}{16} \times 640}$,
and
$\mathbf{f}_3^{u} \in \mathbb{R}^{\frac{H}{8} \times \frac{W}{8} \times 320}$.
This results in a set of multi-scale features $\mathcal{F} = \{\mathbf{f}_k\}_{k=1}^{6}$.

\subsection{Visual Feature injection}
\label{sec:visual_enhancement}

As noted in  \cite{koley2025sketchfusion}, CLIP features tend to be low-frequency (LF) biased, whereas SD features prioritize high-frequency (HF) details.
Motivated by this complementarity, we inject CLIP visual cues into the SD backbone to achieve a more comprehensive representation.
To this end, we extract and integrate both the global class token $\mathbf{c}_{cls}$ and local patch tokens $\mathbf{c}_{patch}$ from CLIP.

\subsubsection{Global Visual Feature Injection}
To incorporate global semantic context, we utilize the CLIP class token $\mathbf{c}_{cls}$.
We adopt the IP-Adapter mechanism to inject this global vector.
Formally, given the class token $\mathbf{c}_{cls}$, we generate $T$ cross-attention tokens via a projection module:
\begin{equation}
\mathbf{t} = \mathrm{ImageProj}(\mathbf{c}_{cls}) \in \mathbb{R}^{T \times d_c}.
\end{equation}
These tokens are injected via the cross-attention mechanism at all timesteps, allowing the UNet to attend to global image-level information alongside textual prompts.

\subsubsection{Local Visual Feature Injection}
While global tokens capture high-level semantics, fine-grained structural details are preserved in CLIP's local patch features.
We extract patch embeddings $\mathbf{c}_{patch}$ from the penultimate layer of the CLIP visual encoder.
To align these features with the SD U-Net, we pass them through a learnable 1D convolutional layer $\mathcal{C}_{inj}(\cdot)$ to modify them just enough to match the feature dimension of different blocks of the UNet.
These transformed feature maps are then added to the intermediate UNet feature maps $\{\mathbf{f}_n^d\}_{n=1}^{3}$ at every timestep of the denoising process:
\begin{equation}
\tilde{\mathbf{f}}_n^d = \mathbf{f}_n^d + \mathcal{C}_{inj}(\mathbf{c}_{patch}).
\end{equation}
This injection occurs in all timesteps and targeted layers of the UNet, enabling the model to influence the denoising process with the dense semantic information encoded within CLIP's visual embeddings.

\subsection{Textual Prompts Injection}
\label{sec:textual_injection}

\subsubsection{BLIP-based Hard Text Prompt Injection}
We employ a pretrained BLIP model to generate textual descriptions from rendered 3D views or sketches.
Specifically, given the visual content $c_v$ and a modality-specific text prefix $c_t$ (e.g., \emph{``a sketch of''} or \emph{``a 3D rendering of''}), the text generation process is formulated as:
\begin{equation}
Text = \text{BLIP}(c_t, c_v).
\end{equation}
This allows the model to auto-complete the caption, resulting in a description that naturally encapsulates both the object category and fine-grained visual details.
While the inferred category may not always strictly align with the ground-truth label, it typically remains proximal within the semantic space.
The generated caption is then processed by the CLIP ViT-L text encoder ($T_L$) within the SDXL framework to produce the hard text embeddings:
\begin{equation}
\mathbf{p}_{\text{hard}} = T_L(Text) \in \mathbb{R}^{m \times d_L},
\end{equation}
where $m=77$ denotes the sequence length and $d_L=768$ is the embedding dimension of the CLIP-L encoder.
These embeddings are subsequently utilized in the cross-attention mechanism to provide explicit semantic guidance.

\subsubsection{Learnable Soft Text Prompt Injection}
Text-to-image diffusion models, particularly SDXL, typically rely on an additional OpenCLIP ViT-bigG (CLIP-G) encoder to extract features for both global and spatial conditioning.
However, standard 3D rendered images and sketch datasets often lack reliable textual annotations.
To address this, we replace the fixed CLIP-G text embeddings with a unified learnable continuous soft prompt tensor:
\begin{equation}
\mathbf{p}_{\text{soft}} \in \mathbb{R}^{(m+1) \times d_G},
\end{equation}
where $d_G=1280$ represents the channel dimension of the CLIP-G encoder.
We utilize the first token of $\mathbf{p}_{\text{soft}}$ to embed into the time embedding module, which modulates the global style of the generation.
Meanwhile, the remaining $m$ tokens of $\mathbf{p}_{\text{soft}}$ are concatenated with the CLIP-L encoded embeddings $\mathbf{p}_{\text{hard}}$ and subsequently fed into the cross-attention layers to control the spatial generation process.

\subsection{SD Feature Aggregation}
\label{sec:feature_aggregation}
Having enriched the multi-scale features with visual and textual guidance, we now proceed to aggregate them into a unified representation.
To map these diverse features into a unified embedding space, we introduce a Feature Adapter module.
For an enhanced feature map $\tilde{\mathbf{f}}_k \in \mathbb{R}^{C_k \times H_k \times W_k}$ from the $k$-th scale, we first apply a 1D convolutional layer $\mathcal{C}_{agg}(\cdot)$ to project it to a target channel dimension $D = 1280$.
The projected features are then processed by a refinement module $R(\cdot)$ composed of three consecutive ResNet blocks.
Finally, we apply global adaptive max pooling to compress the spatial dimensions, resulting in a compact feature vector $\hat{\mathbf{f}}_k \in \mathbb{R}^{D}$.
This process is formulated as:
\begin{equation}
\hat{\mathbf{f}}_k = \operatorname{GMP}\left( R\left( \mathcal{C}_{agg}(\tilde{\mathbf{f}}_k) \right) \right).
\end{equation}

To synthesize the information from all six scales, we aggregate the vectors using a weighted sum with learnable parameters.
We define six learnable weights $\{\alpha_k\}_{k=1}^{6}$, which are normalized via softmax to ensure they sum to one.
The final image-level representation $\bar{\mathbf{f}} \in \mathbb{R}^{D}$ is computed as:
\begin{equation}
\bar{\mathbf{f}} = \sum_{k=1}^{6} w_k \hat{\mathbf{f}}_k, \quad \text{where } w_k = \frac{\exp(\alpha_k)}{\sum_{j=1}^{6} \exp(\alpha_j)}.
\end{equation}

\subsection{Optimization Objectives}
\label{sec:3.5}
\subsubsection{Circle-T Loss: Handling Optimization Imbalance}
\label{sec:Circle-T}

On Sketch-3D datasets, we observe a distinct \textbf{optimization imbalance}: pushing negative samples apart is significantly easier than compacting positive pairs.
Early in training, the model rapidly minimizes loss by separating negatives, resulting in well-separated clusters but weak intra-class compactness.
Inspired by Circle Loss \cite{sun2020circle}, we propose circle-T loss (Targeted-Scaling) to dynamically amplify the focus on positive pairs once negatives are sufficiently distinguished.

Let $\mathbf{F}_a$, $\mathbf{F}_p$, and $\mathcal{N}=\{\mathbf{F}_n\}$ denote the anchor, positive, and negative features, respectively.
We define cosine similarities as $s_p = \mathbf{F}_a^\top \mathbf{F}_p$ and $s_n = \mathbf{F}_a^\top \mathbf{F}_n$.
Unlike the original coupled margin, we decouple the decision boundaries into independent margins $\Delta_p$ and $\Delta_n$.
To achieve targeted optimization, we introduce a Smooth Dynamic Scaling factor $\lambda$, which monitors the average negative similarity $\bar{s}_n = \frac{1}{|\mathcal{N}|}\sum_{\mathbf{F}_n \in \mathcal{N}} (\mathbf{F}_a^\top \mathbf{F}_n)$.
When $\bar{s}_n$ decreases, $\lambda$ increases to impose a heavier penalty on positive pairs:
\begin{equation}
\lambda(\bar{s}_n) = \min \left( 1 + \beta \cdot \exp\left(-\frac{\bar{s}_n}{\tau}\right), \lambda_{\max} \right),
\end{equation}
where $\beta$ and $\tau$ control the scaling magnitude and sensitivity.
The proposed circle-T loss is formulated as:
\begin{equation}
\mathcal{L}_{\text{circle-T}} = \log \left[ 1 + \sum_{\mathbf{F}_n \in \mathcal{N}} \exp \left( \gamma \big( \alpha_n (s_n - \Delta_n) - \lambda(\bar{s}_n) \alpha_p (s_p - \Delta_p) \big) \right) \right],
\end{equation}
where $\gamma$ is the scale factor, and the self-paced weights are defined as $\alpha_p = [2 - \Delta_p - s_p]_+$ and $\alpha_n = [s_n + \Delta_n]_+$.
By amplifying the positive term with $\lambda(\bar{s}_n)$, the loss forces the model to shift gradients toward compacting positive pairs after negative separation.

\subsubsection{View-Specific Classification Loss}
\label{sec:3.5.2}

Complementing metric learning, we introduce a classification constraint.
We observe that while sketch representations naturally possess high semantic separability, 3D view projections ($\mathbf{F}_{\text{view}}$) often share geometric similarities across categories, making them harder to separate.
To mitigate this, we impose a Cross-Entropy loss $\mathcal{L}_{\text{cls}}^{\text{view}}$ exclusively on the view branch to enforce explicit class boundaries:
\begin{equation}
\mathcal{L}_{\text{cls}}^{\text{view}} = - \frac{1}{N} \sum_{i=1}^{N} \log \frac{\exp(\mathbf{W}_{y_i}^\top \mathbf{F}_{i,\text{view}})}{\sum_{j=1}^{C} \exp(\mathbf{W}_j^\top \mathbf{F}_{i,\text{view}})}.
\end{equation}

\subsubsection{Total Objective}
\label{sec:3.5.3}

The final objective aggregates the circle-T loss from both sketch-anchored ($\mathcal{L}_{\text{circle-T}}^{\text{ske}}$) and view-anchored ($\mathcal{L}_{\text{circle-T}}^{\text{view}}$) branches, along with the view classification loss:
\begin{equation}
\mathcal{L}_{\text{total}} = \mathcal{L}_{\text{circle-T}}^{\text{ske}} + \mathcal{L}_{\text{circle-T}}^{\text{view}} + \eta \cdot \mathcal{L}_{\text{cls}}^{\text{view}},
\end{equation}
where $\eta$ balances the metric and classification objectives.

\section{Experiments}
\subsection{Experiment Setup}
\begin{table*}[t]
  \caption{Comparison with state-of-the-art methods on Zero-shot SBSR for benchmark SHREC2013 and SHREC2014 with split I.}
  \label{tab:main_shrec13}
  \centering
  \setlength{\tabcolsep}{2.5pt}
  \begin{tabularx}{\textwidth}{c c c *{7}{Y} *{7}{Y}}
    \toprule
    \multirow{2}{*}{Type} & \multirow{2}{*}{Method} & \multirow{2}{*}{Publication}
    & \multicolumn{7}{c}{SHREC’2013}
    & \multicolumn{7}{c}{SHREC’2014} \\
    \cmidrule(lr){4-10} \cmidrule(lr){11-17}
    & & 
    & NN$\uparrow$ & FT$\uparrow$ & ST/2$\uparrow$ & nDCG$\uparrow$ & \,1-E$\downarrow$ & MRR$\uparrow$ & mAP$\uparrow$
    & NN$\uparrow$ & FT$\uparrow$ & ST/2$\uparrow$ & nDCG$\uparrow$ & \,1-E$\downarrow$ & MRR$\uparrow$ & mAP$\uparrow$ \\
    \midrule
    & B-DINO \cite{caron2021DINO} & ICCV2021
    & 6.6 & 9.7 & 9.1 & 39.5 & 82.8 & 18.0 & 15.7
    & 7.4 & 7.3 & 6.8 & 34.2 & 84.0 & 15.4 & 13.7 \\
    & B-SD \cite{rombach2022high} & CVPR2022
    & 5.1 & 5.0 & 4.7 & 34.5 & 88.5 & 13.4 & 9.8
    & 7.2 & 5.5 & 5.9 & 30.3 & 88.8 & 13.7 & 9.5 \\
    & B-SD-CLIP & --
    & 5.0 & 6.6 & 6.3 & 37.2 & 85.7 & 16.1 & 12.1
    & 15.9 & 12.0 & 12.2 & 37.4 & 82.2 & 24.3 & 16.7 \\
    \midrule
    \multirow{7}{*}{View Enc}
    & Siamese \cite{Wang2015CVPR} & CVPR2015
    & 13.7 & 12.4 & 11.7 & 18.6 & 76.2 & 25.9 & 20.1
    & 10.0 & 8.6 & 8.7 & 13.6 & 78.4 & 21.4 & 17.0 \\
    & DCHML \cite{Dai2018TIP} & TIP2018
    & 18.3 & 21.1 & 13.6 & 23.8 & 63.2 & 27.8 & 33.9
    & 16.7 & 17.7 & 17.9 & 27.3 & 67.3 & 27.8 & 30.0 \\
    & TCL \cite{he2018triplet} & CVPR2018
    & 25.0 & 17.8 & 20.3 & 31.4 & 65.0 & 39.0 & 31.7
    & 21.3 & 21.6 & 19.1 & 30.6 & 66.5 & 31.4 & 32.8 \\
    & CGN \cite{Dai2020ICME} & ICME2020
    & 33.3 & 29.4 & 22.5 & 39.8 & 63.4 & 48.4 & 38.6
    & 26.3 & 25.0 & 21.4 & 35.1 & 65.0 & 36.8 & 35.6 \\
    & PCL \cite{Wang2023AAAI} & AAAI2023
    & \underline{38.3} & \underline{38.9} & \underline{26.1} & \underline{47.0} & \textbf{54.9} & \underline{52.4} & \underline{48.0}
    & \underline{33.3} & \underline{32.8} & \underline{22.8} & \underline{39.9} & \underline{57.8} & \underline{45.5} & \underline{43.4} \\
    & CFTTSL \cite{Liang2024CVIU} & CVIU2024
    & 34.9 & 26.1 & 19.1 & 28.1 & 70.4 & 46.6 & 32.5
    & 16.7 & 14.4 & 11.3 & 35.8 & 76.7 & 26.6 & 21.6 \\
    & MEHA \cite{zhang2025meha} & SIGIR2025
    & 33.9 & 27.7 & {22.3} & 46.9 & 70.5 & 47.8 & 34.9
    & 10.8 & 10.4 & 15.8 & 30.7 & 89.4 & 20.1 & 15.3 \\
    \midrule
    & \textbf{Diff-SBSR(Ours)} &  
    & \textbf{48.9} & \textbf{40.9} & \textbf{31.3} & \textbf{57.2} & \underline{61.9} & \textbf{68.5} & \textbf{49.9} 
    & \textbf{48.3} & \textbf{40.1} & \textbf{28.3} & \textbf{65.4} & \textbf{53.2} & \textbf{59.3} & \textbf{49.9} \\
    \bottomrule
  \end{tabularx}
\end{table*}

\begin{table*}[t]
  \caption{Comparison with state-of-the-art methods on Zero-shot SBSR for benchmark SHREC2013 and SHREC2014 with split II.}
  \label{tab:main_shrec14}
  \centering
  \setlength{\tabcolsep}{2.5pt}
  \begin{tabularx}{\textwidth}{c c c *{6}{Y} *{6}{Y}}
    \toprule
    \multirow{2}{*}{Type} & \multirow{2}{*}{Method} & \multirow{2}{*}{Publication}
    & \multicolumn{6}{c}{SHREC2013}
    & \multicolumn{6}{c}{SHREC2014} \\
    \cmidrule(lr){4-9} \cmidrule(lr){10-15}
    & & 
    & NN$\uparrow$ & FT$\uparrow$ & ST$\uparrow$ & E$\uparrow$ & DCG$\uparrow$ & mAP$\uparrow$
    & NN$\uparrow$ & FT$\uparrow$ & ST$\uparrow$ & E$\uparrow$ & DCG$\uparrow$ & mAP$\uparrow$ \\
    \midrule
    & B-DINO \cite{caron2021DINO} & ICCV2021
    & 15.2 & 10.5 & 18.4 & 16.6 & 45.0 & 17.3
    & 11.1 & 8.4 & 16.1 & 4.9 & 42.6 & 10.5 \\
    & B-SD \cite{rombach2022high} & CVPR2022
    & 20.7 & 8.8 & 19.2 & 14.6 & 46.9 & 15.8
    & 5.7 & 5.3 & 9.5 & 2.8 & 38.4 & 7.4 \\
    & B-SD-CLIP & --
    & 34.8 & 17.8 & 29.1 & 17.2 & 53.4 & 22.7
    & 15.9 & 12.0 & 18.4 & 8.5 & 37.4 & 16.7 \\
    \midrule
    \multirow{7}{*}{View Enc}
    & Siamese \cite{Wang2015CVPR} & CVPR2015
    & 13.7 & 11.4 & 20.3 & 16.2 & 40.4 & 17.1
    & 9.7 & 10.2 & 11.3 & 5.2 & 31.4 & 10.8 \\
    & DCHML \cite{Dai2018TIP} & TIP2018
    & 31.8 & 30.4 & 42.1 & 28.8 & 58.1 & 36.1
    & 15.7 & 13.4 & 14.5 & 8.4 & 37.9 & 18.7 \\
    & TCL \cite{he2018triplet} & CVPR2018
    & 33.7 & 35.7 & 53.7 & 27.8 & 58.9 & 42.6
    & 27.9 & 25.7 & 15.3 & 12.5 & 45.9 & 23.7 \\
    & CGN \cite{Dai2020ICME} & ICME2020
    & 51.2 & 45.8 & 64.7 & 34.7 & 67.3 & 51.5
    & 40.1 & 32.4 & 42.9 & 17.8 & 57.1 & 33.2 \\
    & DD-GAN \cite{Xu2022AAAI} & AAAI2022
    & 54.4 & 48.4 & 66.1 & 36.4 & 69.6 & 55.1
    & 42.5 & 35.4 & 46.2 & 19.6 & 59.2 & 37.1 \\
    & Pivoting \cite{Chowdhury2023ICCV} & ICCV2023
    & 55.3 & 49.5 & 67.5 & 37.4 & 70.1 & 55.7 
    & 43.1 & 35.9 & 46.8 & 19.6 & 59.5 & 37.6 \\
    & Codi \cite{Meng2024TCSVT} & TCSVT2025
    & \underline{58.7} & \underline{58.6} & \underline{70.6} & \underline{39.5} & \underline{74.2} & \textbf{63.7}
    & \underline{54.0} & \underline{37.2} & \underline{49.5} & \underline{24.5} & \underline{64.2} & \underline{39.5} \\
    \midrule

    & \textbf{Diff-SBSR(Ours)} &  
    & \textbf{61.7}& \textbf{59.2}& \textbf{78.3}& \textbf{40.3}& \textbf{79.9}& \underline{63.5} 
    & \textbf{59.3}& \textbf{40.8}& \textbf{51.9}& \textbf{33.5}& \textbf{68.3}& \textbf{43.8}\\
    \bottomrule
  \end{tabularx}
\end{table*}

\subsubsection{Datasets}
We evaluate the proposed framework on two widely used benchmarks for sketch-based 3D shape retrieval, SHREC2013 \cite{Li2013SHREC} and SHREC2014 \cite{Li2014SHREC}. SHREC2013 consists of 1,258 3D shapes and 7,200 freehand sketches from 90 categories, with 80 sketches per category (50 for training and 30 for testing). SHREC2014 follows the same protocol but is larger and more challenging, containing 8,987 3D shapes and 13,680 sketches across 171 categories. Owing to higher inter-class similarity and greater intra-class variation, SHREC2014 presents a more demanding benchmark. For zero-shot evaluation, we adopt two standard data splits. Split-I follows the protocol in \cite{Xu2022AAAI}, where categories are divided alphabetically: for SHREC2013, 79 categories are used for training and 11 for testing, while for SHREC2014, 151 categories are used for training and 20 for testing. Split-II follows \cite{Wang2023AAAI} and defines unseen categories based on data scarcity. In SHREC2013, 23 categories with five or fewer 3D shapes are treated as unseen, with the remaining 67 used for training; in SHREC2014, 38 such categories are designated as unseen, and the remaining 133 categories form the training set.

\subsubsection{Evaluation Metrics}
We adopt a comprehensive set of standard retrieval metrics, including nearest neighbor (NN), first-tier (FT), second-tier (ST), discounted cumulative gain (DCG), normalized DCG (nDCG), E-measure (E), $1-E$, mean reciprocal rank (MRR), mean average precision (mAP), and an alternative sketch fidelity metric (SF). NN measures the percentage of queries with the top-1 retrieval in the correct category. FT and ST compute the fraction of correctly labeled 3D shapes within the top $(C-1)$ and $2(C-1)$ results, where $C$ is the number of 3D shapes per category. E integrates precision and recall, with $1-E$ complementing it to emphasize retrieval error \cite{Wang2023AAAI}. DCG weights correct results by rank, and nDCG normalizes this by the ideal ranking. MRR evaluates the reciprocal rank of the first relevant retrieval, while mAP averages precision over all queries.

\subsubsection{Implementation Details}
In all experiments, we adopt Stable Diffusion XL as our backbone, where the CLIP embedding dimension is set to $d=1024$. Both the U-Net denoiser and the VAE encoder are kept frozen throughout training. We optimize the learnable prompts for 100 epochs using the AdamW optimizer with a weight decay of 0.09, a batch size of 50, and a learning rate of $1\times10^{-4}$ on an NVIDIA GTX 4090 GPU.

\subsection{Comparison to State-of-the-Art Methods}
We compare our method with several state-of-the-art baselines, including Siamese \cite{Wang2015CVPR}, DCHML \cite{Dai2018TIP}, TCL \cite{he2018triplet}, CGN \cite{Dai2020ICME}, PCL \cite{Wang2023AAAI}, CFTTSL \cite{Liang2024CVIU}, MEHA \cite{zhang2025meha}, Pivoting \cite{Chowdhury2023ICCV}, Codi \cite{Meng2024TCSVT}. We also include results from recent diffusion-based baselines, namely B-DINO \cite{caron2021DINO}, B-SD \cite{rombach2022high}, and B-SD-CLIP (our re-implementation of B-SD with CLIP visual features injected). For a fair comparison, all methods are evaluated under the same zero-shot protocols without any fine-tuning on unseen categories. Tables~\ref{tab:main_shrec13} and \ref{tab:main_shrec14} report quantitative comparisons with state-of-the-art methods on SHREC2013 and SHREC2014 under the two standard zero-shot evaluation protocols. 

\noindent\textbf{Results on Split-I}
As shown in Table~\ref{tab:main_shrec13}, Diff-SBSR consistently outperforms all competing methods across most evaluation metrics on both SHREC2013 and SHREC2014 under Split-I. In particular, our method achieves substantial improvements on NN, ST/2, nDCG, MRR, and mAP, indicating more accurate nearest-neighbor retrieval, improved ranking quality, and stronger overall retrieval performance in zero-shot settings. On SHREC2013, Diff-SBSR surpasses the strongest prior method PCL by a clear margin in NN (+10.6\%), nDCG (+10.2\%), and MRR (+16.1\%), demonstrating the effectiveness of diffusion-based representations in capturing geometry-consistent features under sketch abstraction. On the more challenging SHREC2014 benchmark, Diff-SBSR achieves even larger gains, particularly in NN and nDCG, reflecting its improved robustness to higher inter-class similarity and intra-class variation. While Diff-SBSR does not achieve the lowest value on the 1–E metric for SHREC2013, the difference is relatively small compared to PCL. Notably, our method attains consistently higher performance on ranking-based metrics such as MRR and mAP, suggesting that it produces more reliable global rankings despite minor variations in early retrieval errors.

\noindent\textbf{Results on Split-II}
Table~\ref{tab:main_shrec14} reports the comparison results under the more challenging Split-II protocol, where unseen categories are defined based on data scarcity. Under this setting, Diff-SBSR again achieves the best overall performance on the majority of metrics across both datasets. On SHREC2013, Diff-SBSR obtains the highest scores on NN, ST, DCG, and mAP, indicating strong generalization to categories with extremely limited 3D samples. On SHREC2014, our method consistently outperforms prior approaches such as DD-GAN, Pivoting, and Codi, particularly on NN and DCG, which are critical for evaluating retrieval accuracy and ranking quality in zero-shot scenarios. For certain metrics, such as FT on SHREC2013, Diff-SBSR performs comparably to but does not exceed the best-performing baseline. This behavior can be attributed to the increased emphasis of Diff-SBSR on global semantic consistency and geometry-aware feature alignment, which primarily benefits overall ranking quality rather than fine-grained retrieval at very shallow depths. Nevertheless, the consistent improvements across NN, DCG, and mAP demonstrate the overall superiority of Diff-SBSR in zero-shot SBSR.

\subsection{Ablation Study}
\subsubsection{The Influence of Components}
We evaluate the contribution of each component in our feature enhancement module on SHREC2013 (Table~\ref{tab:Ablation}). Removing the global visual prompt reduces NN from 63.5\% to 58.5\% and mAP from 62.7\% to 51.5\%, highlighting the importance of CLIP-injected global semantics for category-level alignment. Disabling the local visual prompt causes a sharper NN drop to 48.2, indicating that patch-level cues are critical for fine-grained shape details. For textual conditioning, removing soft prompts mainly affects ranking metrics (mAP -5.9\%, DCG -4.6\%), while removing hard prompts yields larger drops across most metrics (NN -9.3\%, ST -16.2\%), emphasizing the role of BLIP-generated descriptions. The consistent performance degradation from ablating any component confirms that multimodal semantic priors are essential for producing discriminative and robust representations for zero-shot sketch-based 3D shape retrieval.

\begin{table}[h]
  \centering
  \caption{Ablation on SHREC2013 with Components on SHREC2013 dataset with split II.}
  \label{tab:Ablation}
  \begin{tabularx}{\linewidth}{>{\centering\arraybackslash}p{3.3cm}*{7}{>{\centering\arraybackslash}X}}
    \toprule
    Components & NN & FT & ST & E & DCG & mAP \\
    \midrule
    w/o global visual prompt
    & 58.5 & 48.1 & 64.1 & 33.6 & 73.4 & 51.5 \\
    w/o local visual prompt
    & 48.2 & 44.7 & 66.5 & 37.0 & 71.9 & 49.4 \\
    w/o hard text prompt
    & 54.2 & 43.5 & 62.7 & 34.8 & 72.1 & 49.5 \\
    w/o soft text prompt
    & 56.4 & 51.7 & 69.3 & 35.8 & 75.2 & 56.8 \\    
    w/o l2 of soft text prompt
    & 60.0 & 46.1 & 58.2 & 30.6 & 71.8 & 50.2 \\
    \rowcolor{lightpurple}
    FULL
    & \textbf{61.7}& \textbf{59.2}& \textbf{78.3}& \textbf{40.3}& \textbf{79.9}& \textbf{63.5}  \\
    \bottomrule
  \end{tabularx}
\end{table}

\subsubsection{The Influence of Learning Objective for Alignment}

Table~\ref{tab:Loss_ablation} analyzes the effect of different loss combinations for cross-modal matching on SHREC2013 (Split II). Removing either view-level or sketch-level circle loss consistently degrades performance, indicating that symmetric metric learning on both modalities is essential for aligning sketch and 3D representations in the shared embedding space. In particular, excluding $\mathcal{L}_{circle}^{view}$ or $\mathcal{L}_{circle}^{ske}$ leads to noticeable drops in NN and mAP, suggesting weakened cross-modal discriminability. Removing the view-level classification loss $\mathcal{L}_{cls}^{view}$ leads to a slight performance drop, indicating that category labels for rendered views provide complementary semantic supervision that facilitates representation learning. In contrast, removing the sketch-level classification loss $\mathcal{L}_{cls}^{ske}$ results in a consistent performance improvement, achieving the best overall results. This suggests that sketches inherently exhibit high noise and sparsity, and enforcing hard class supervision may induce overfitting, thereby weakening zero-shot generalization. In this case, metric-based supervision offers a more flexible and robust learning signal. Overall, the results demonstrate that carefully balancing symmetric circle losses with selective classification supervision is crucial for robust cross-modal retrieval, and excessive constraints may hinder effective representation learning.

\begin{table}[h]
\centering
\caption{Influence of each loss function for cross-model matching on SHREC2013 dataset with split II.}
\label{tab:Loss_ablation}
\setlength{\tabcolsep}{4pt}
\begin{tabularx}{\columnwidth}{p{0.6cm} p{0.6cm} p{0.6cm} p{0.6cm} *{6}{Y}}
\toprule
\multicolumn{4}{c}{Objective} & \multicolumn{6}{c}{SHREC 2013} \\
\midrule
$\mathcal{L}_{circle}^{view}$ & $\mathcal{L}_{circle}^{ske}$ & $\mathcal{L}_{cls}^{view}$ & $\mathcal{L}_{cls}^{ske}$ & NN & FT & ST & E & DCG & mAP \\
\midrule
$\times$ & $\checkmark$ & $\checkmark$ & $\checkmark$ & 53.6 & 48.4 & 65.3 & 37.0 & 73.0 & 54.0 \\
$\checkmark$ & $\times$ & $\checkmark$ & $\checkmark$ & 53.2 & 49.4 & 64.4 & 38.9 & 73.9 & 55.1 \\
$\checkmark$ & $\checkmark$ & $\times$ & $\checkmark$ & 65.7 & 53.6 & 70.3 & 38.1 & 77.9 & 60.2 \\
\rowcolor{lightpurple}
$\checkmark$ & $\checkmark$ & $\checkmark$ & $\times$ & \textbf{61.7} & \textbf{59.2} & \textbf{78.3} & \textbf{40.3} & \textbf{79.9} & \textbf{63.5} \\
$\checkmark$ & $\checkmark$ & $\checkmark$ & $\checkmark$ & 59.6 & 48.1 & 64.0 & 35.3 & 74.6 & 53.0 \\
\bottomrule
\end{tabularx}
\end{table}

\subsubsection{The Influence of values of $\eta$ for $\mathcal{L}_{cls}$}
Table \ref{tab:values} reports the impact of varying $\eta$. With small weights ($\eta=0.1$ or $1$), performance is consistently lower (e.g., mAP drops to 50.7 and 46.7), indicating that weak classification supervision provides insufficient semantic regularization for view embeddings. Increasing $\eta$ to 10 yields the best results across all metrics, improving NN from 51.8 to 63.5 and mAP from 50.7 to 62.7, which shows that moderate view-level supervision effectively complements metric learning by reinforcing category structure. When $\eta$ is further increased to 100, performance degrades again (mAP 49.7), suggesting that overly strong classification loss biases the model toward closed-set discrimination and harms cross-modal alignment. Overall, $\mathcal{L}_{cls}^{view}$ acts as a regularizer, and a balanced weighting is essential for optimal zero-shot retrieval.

\begin{table}[h]
  \centering
  \caption{The Influence of values of $\eta$ for $\mathcal{L}_{cls}^{view}$.}
  \label{tab:values}
  \begin{tabularx}{\linewidth}{>{\centering\arraybackslash}p{1.5cm} * {7}{>{\centering\arraybackslash}X}}
    \toprule
    $\eta$ & NN & FT & ST & E & DCG & mAP \\
    \midrule
    0.1  & 51.8 & 45.9 & 62.1 & 34.8 & 71.6 & 50.7 \\
    1 & 48.6 & 41.8 & 57.8 & 32.5 & 68.0 & 46.7 \\
    \rowcolor{lightpurple}
    10 & \textbf{61.7} & \textbf{59.2} & \textbf{78.3} & \textbf{40.3} & \textbf{79.9} & \textbf{63.5} \\
    100 & 60.1 & 43.2 & 62.3 & 36.5 & 72.6 & 49.7 \\
    \bottomrule
  \end{tabularx}
\end{table}

\subsubsection{Influence of Different Timesteps}
\begin{figure}[h]
  \centering
  \includegraphics[
    width=0.5\linewidth,
    trim=0mm 0mm 0mm 0mm,
    clip
]{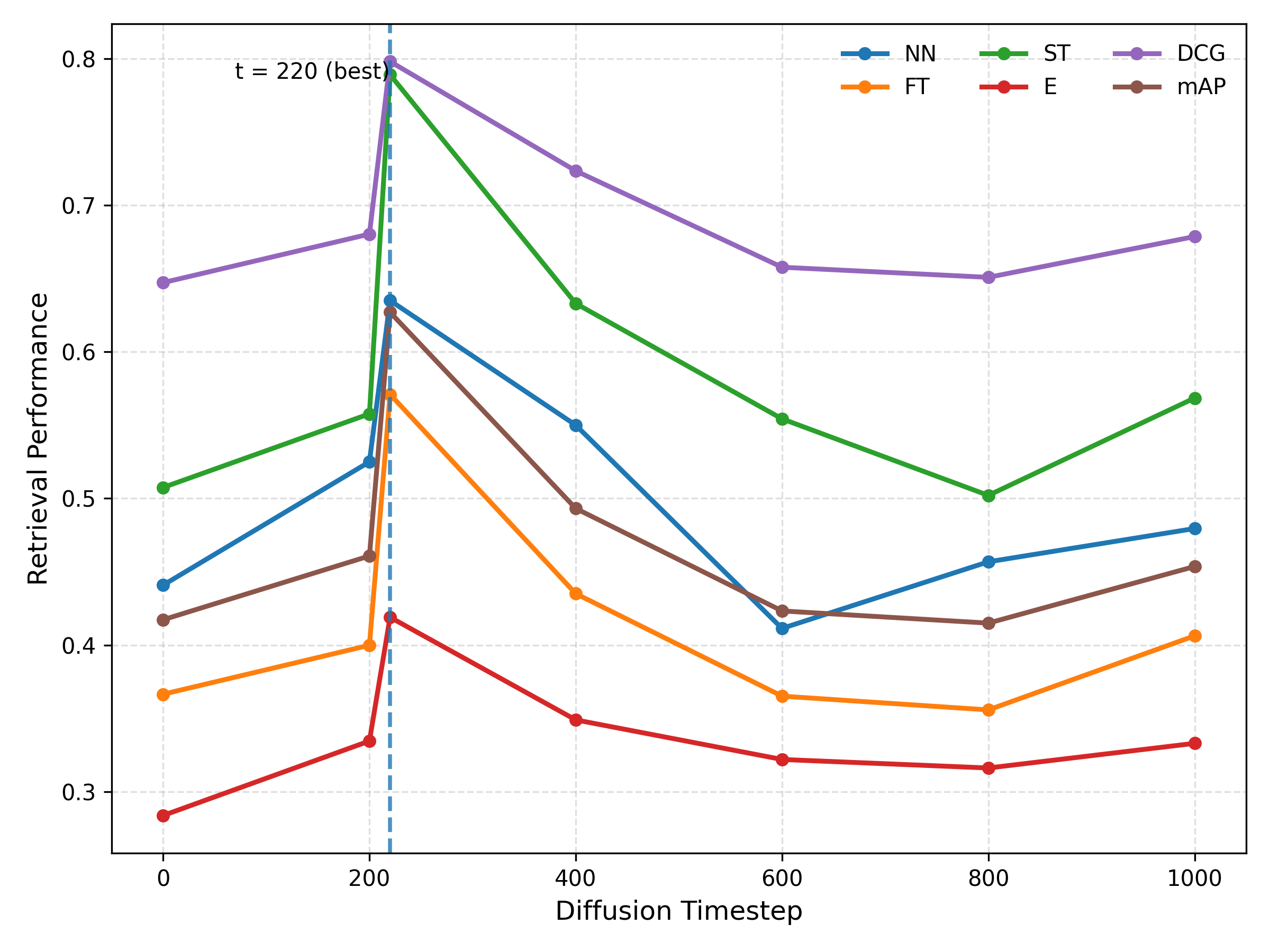}
  \caption{Effect of timesteps choice of diffusion model on the SHREC2013 dataset on SHREC2013 dataset with split II.}
  \Description{Compared with different diffusion timesteps, showing how feature clusters evolve across timesteps.}
  \label{figure:Timestep}
\end{figure}

We investigate the impact of diffusion timesteps on feature quality by extracting representations from Stable Diffusion at $t={0,200,220,400,600,800,1000}$ and evaluating their retrieval performance. As shown in Fig.~5, different timesteps yield markedly different feature characteristics, leading to varying retrieval accuracy. Nevertheless, our method exhibits strong robustness to timestep selection: across a wide range of $t$, the performance consistently surpasses the baseline, indicating that our approach does not rely on a narrowly tuned timestep. Among them, $t=220$ achieves the best performance and is therefore adopted as our default setting in all experiments (Fig.~\ref{figure:Timestep}).

\subsection{Visualization}
\subsubsection{PCA Visualization of Feature-Enhanced Representations.}
\begin{figure}[t]
  \centering
  \includegraphics[
    width=\linewidth,
    trim=0mm 0mm 0mm 62mm,
    clip
]{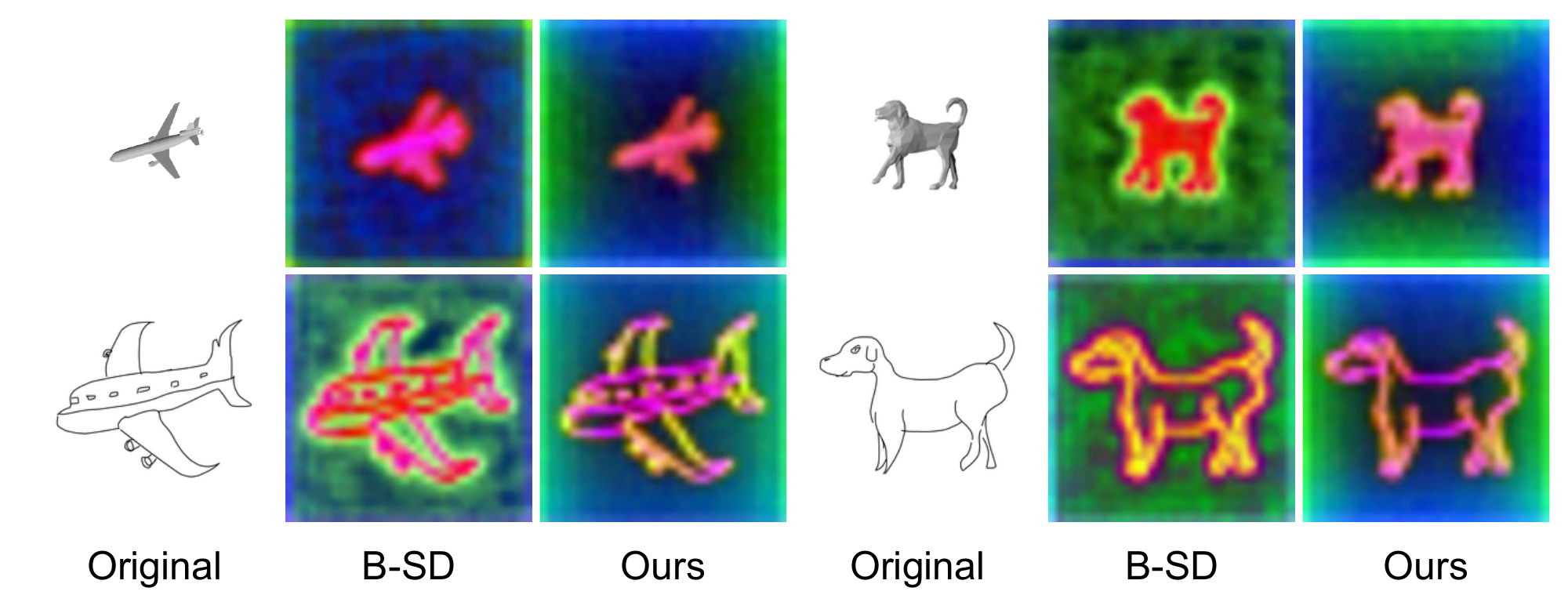}
  \caption{PCA representation of SD’s intermediate UNet features on B-SD and our method.}
  \label{fig:pca}
\end{figure}

We perform PCA on intermediate U-Net features of Stable Diffusion to analyze their semantic structure (Fig.~\ref{fig:pca}). In the B-SD baseline, sketch features produce scattered background noise (green), reflecting SD’s bias toward high-frequency signals. Pretrained on natural images, the U-Net skip connections propagate edge- and texture-related responses, which on sparse binary sketches leads to indiscriminate activations in blank areas. In contrast, our feature-enhanced approach injects explicit semantic constraints: local CLIP visual features provide stable structural priors, and global visual–textual conditioning suppresses irrelevant background responses. As a result, background noise is largely removed, and high-response components (purple–red) concentrate along sketch contours, achieving clearer foreground–background separation and structure-aligned representations for sketch-based cross-modal retrieval.

\subsubsection{T-SNE Visualization of Feature-Enhanced Representations.}
\begin{figure}[t]
  \centering
  \includegraphics[
    width=\linewidth,
    trim=0mm 0mm 0mm 64mm,
    clip
]{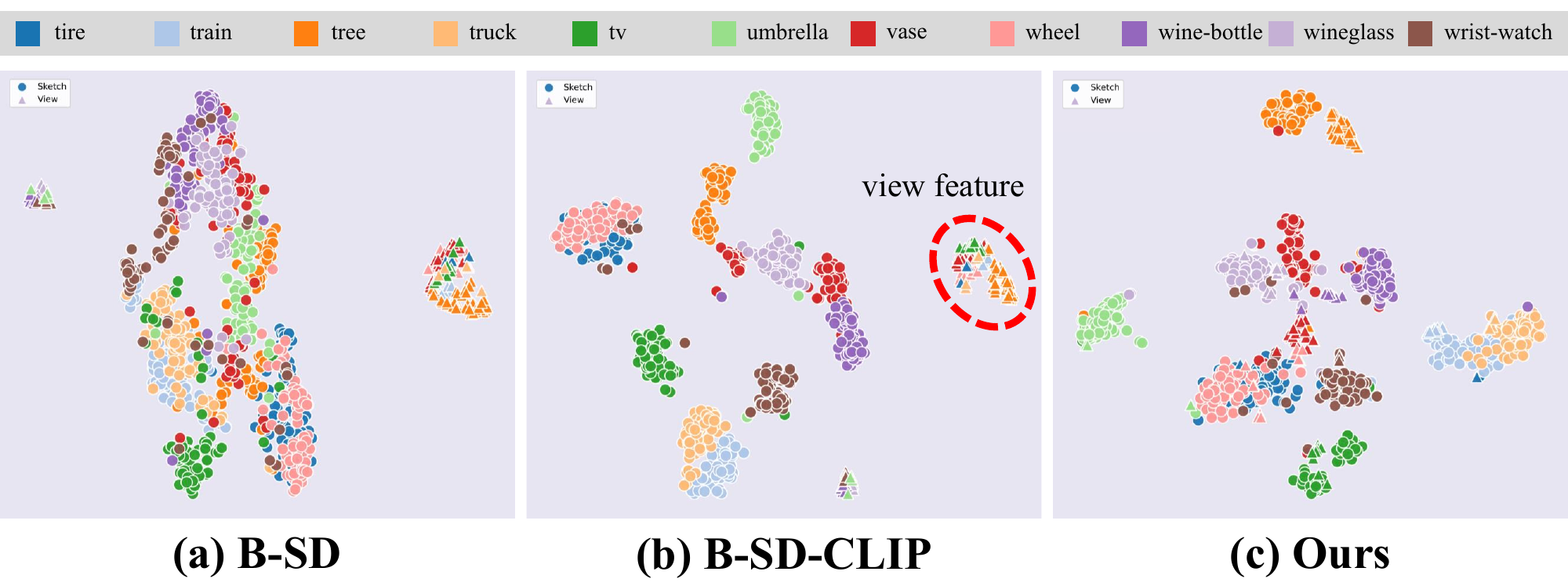}
  \caption{t-SNE visualization of feature representations on SHREC2013 under Split-II: (a) baseline Stable Diffusion (B-SD); (b) our feature-enhanced method.}
  \label{fig:tsne}
\end{figure}

To intuitively evaluate the quality of the learned embedding space, we visualize the feature distributions of the retrieval results on the SHREC13 dataset using t-SNE.
In the B-SD baseline (Fig.~\ref{fig:tsne}(a)), we observe a significant modality gap between sketches and 3D view images. Furthermore, the inter-class discriminability among sketches is poor, primarily due to the inherent sparsity and abstraction of sketch data.
Upon incorporating CLIP features in the B-SD-CLIP baseline (Fig.~\ref{fig:tsne}(b)), we observe that while the cross-modal gap remains substantial and difficult to bridge, the separability between different sketch categories improves significantly. This improvement can be attributed to the complementary nature of the features: while SD prioritizes high-frequency structural details, CLIP supplements the global semantic and low-frequency information that is often lacking in pure SD representations.
Finally, with our proposed method (Fig.~\ref{fig:tsne}(c)), leveraged by the visual-textual enhancement strategy, we effectively bridge the modality gap between sketches and projected views. The resulting feature space exhibits clear inter-class separation and compact intra-class clustering, demonstrating superior retrieval performance.

\subsubsection{Similarity Heatmaps with circle-T loss and triplet loss.}
\begin{figure}[h]
  \centering
  \includegraphics[
    width=\linewidth,
    trim=0mm 0mm 0mm 27mm,
    clip
]{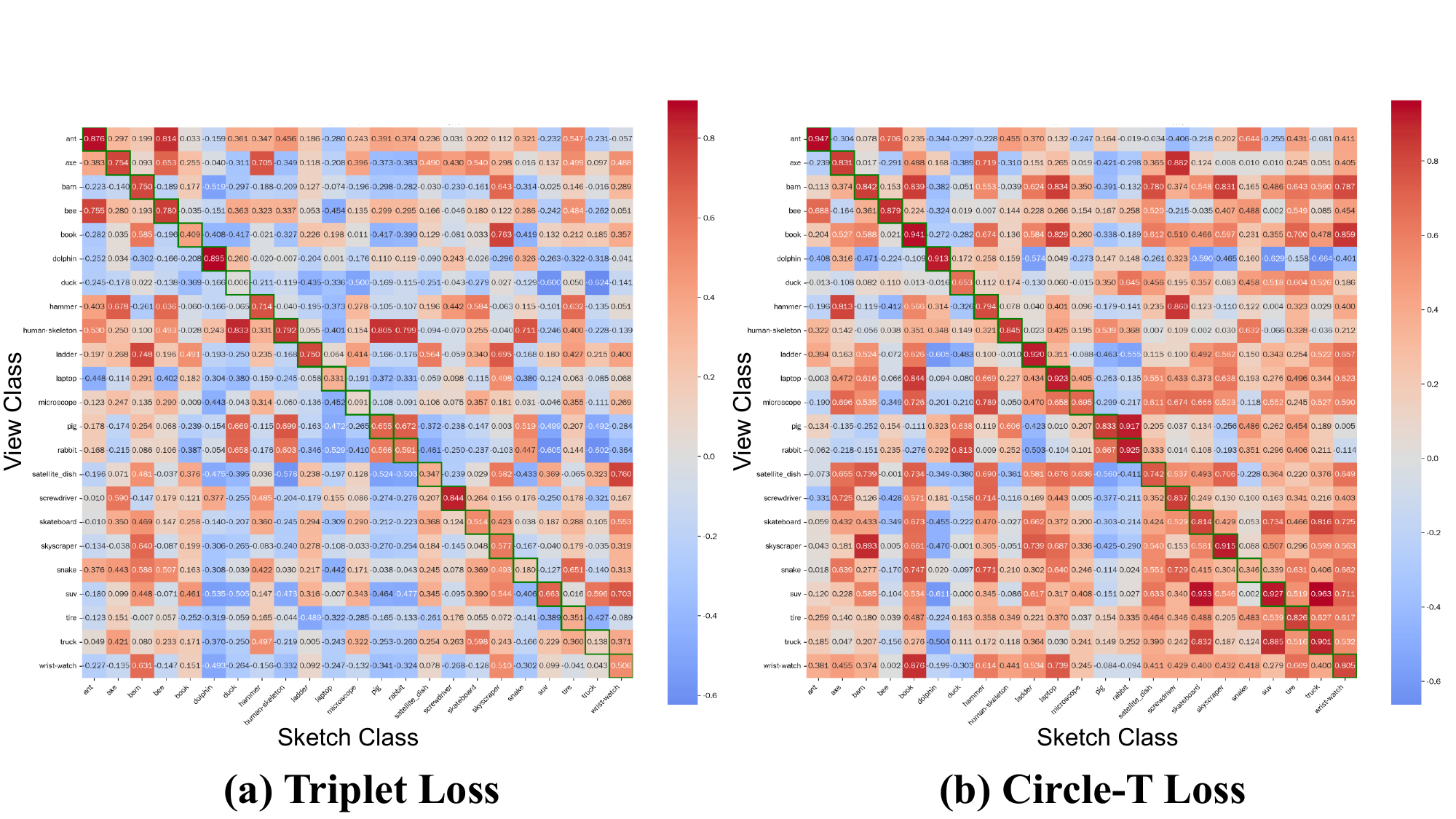}
    \caption{Confusion matrix of retrieval results under circle-T loss and triplet loss on SHREC2013 dataset with split I}
    \label{fig:heatmap}
\end{figure}

We visualize the retrieval consistency between sketch and 3D categories using class-wise similarity heatmaps in Fig. ~\ref{fig:heatmap}, where stronger red responses indicate better cross-modal alignment. Compared with Triplet Loss, Circle Loss produces a much sharper and more concentrated diagonal pattern, while significantly suppressing off-diagonal activations. This indicates stronger intra-class compactness and reduced cross-category confusion. The advantage of Circle Loss stems from its unified reweighting of all positive and negative pairs in an angular space, which simultaneously enlarges inter-class margins and tightens intra-class distributions, without relying on hard sample mining. As a result, it yields a more stable and globally structured embedding space, leading to cleaner category-level alignment between sketches and 3D shapes.

\subsubsection{Visualization of Retrieval Examples}
\begin{figure}[t]
  \centering

  \includegraphics[
    width=0.9\linewidth,
    trim=0mm 0mm 0mm 0mm,
    clip
]{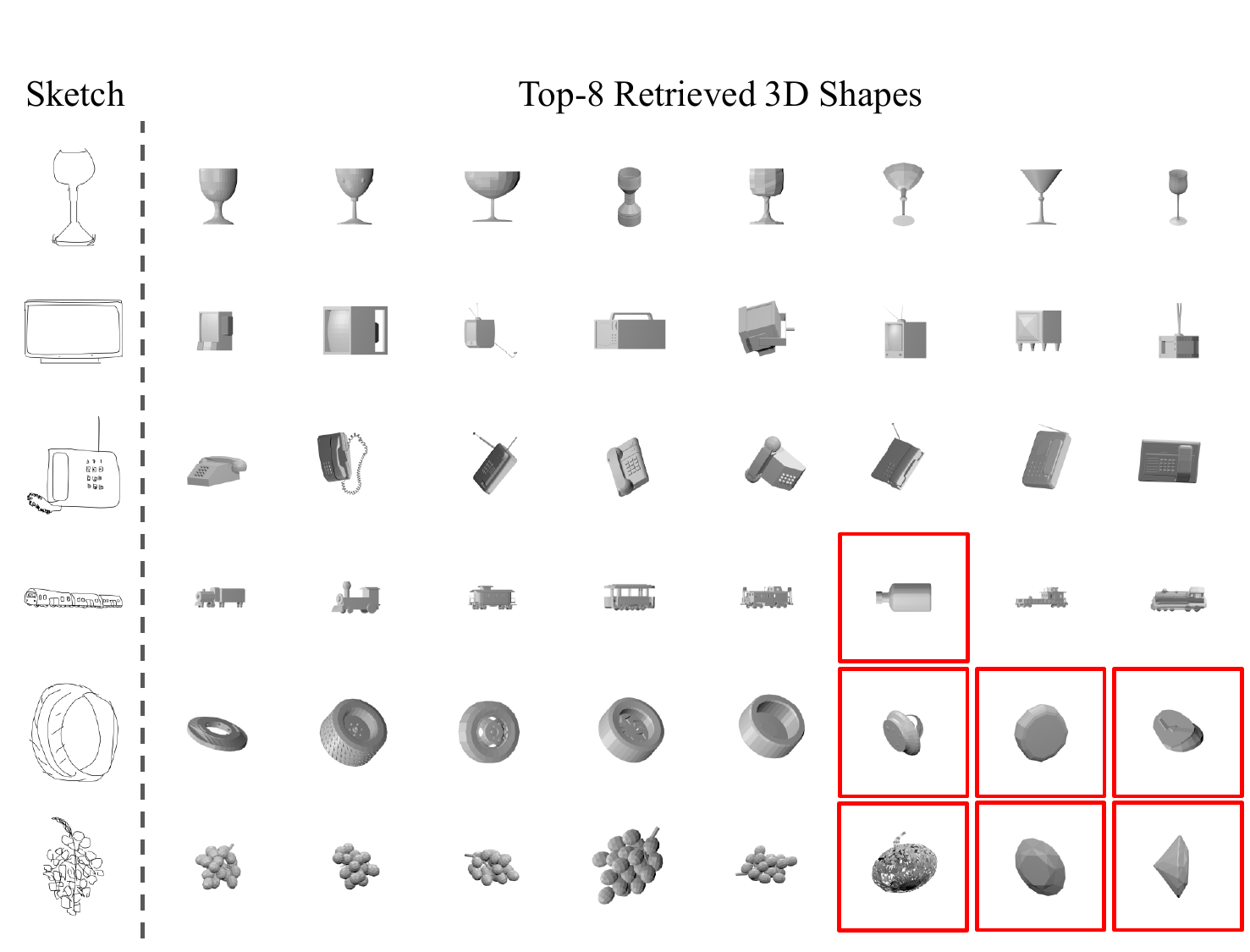}
  \caption{Visualization of retrieval examplesl on SHREC2013 and SHREC2014 dataset.}
  \label{fig:result}
\end{figure}

We visualize representative query sketches and their retrieved 3D shapes in Fig.~\ref{fig:result}. Our method reliably returns correct results within the top few ranks for most queries, indicating strong cross-modal alignment and robust zero-shot generalization despite variations in viewpoint and sketch sparsity. We observe occasional failures, which we mainly attribute to severe class imbalance and the scarcity of positive samples per category, limiting the model’s ability to capture full intra-class diversity. Overall, the qualitative results demonstrate the effectiveness of our framework for practical sketch-based 3D retrieval.

\section{Conclusion}

In this work, we have presented a novel framework for zero-shot sketch-based 3D shape retrieval that leverages the rich intermediate representations of pretrained Stable Diffusion models while explicitly addressing their limitations in sparse, abstract sketch domains. By integrating a CLIP visual encoder and carefully designed multi-scale, multi-modal feature injection, our approach effectively fuses high-frequency structural cues with low-frequency semantic information, producing discriminative and semantically aligned embeddings for both sketches and 3D views. Extensive experiments demonstrate that our framework consistently outperforms existing baselines in zero-shot settings, highlighting the robustness of diffusion-based features when complemented with semantic priors. Beyond immediate performance gains, our analysis provides insights into the role of timestep selection, layer-wise feature aggregation, and hybrid semantic reinforcement, establishing practical guidelines for adapting generative models as feature extractors in cross-modal retrieval tasks. These results underscore the potential of combining large-scale pretrained generative models with task-specific adaptation to bridge human abstraction and machine perception in challenging zero-shot retrieval scenarios.


\bibliographystyle{ACM-Reference-Format}
\bibliography{sample-base}

\end{document}